\documentclass[sigconf]{acmart}
\AtBeginDocument{%
 \providecommand\BibTeX{{%
 \normalfont B\kern-0.5em{\scshape i\kern-0.25em b}\kern-0.8em\TeX}}}

\copyrightyear{2021} 
\acmYear{2021} 
\setcopyright{acmcopyright}\acmConference[MM '21]{Proceedings of the 29th ACM International Conference on Multimedia}{October 20--24, 2021}{Virtual Event, China}
\acmBooktitle{Proceedings of the 29th ACM International Conference on Multimedia (MM '21), October 20--24, 2021, Virtual Event, China}
\acmPrice{15.00}
\acmDOI{10.1145/3474085.3475693}
\acmISBN{978-1-4503-8651-7/21/10}

\begin{document}
\fancyhead{}

%% Use this when submitting an article to a sponsored event. You'll
%% receive a unique submission ID from the organizers
%% of the event, and this ID should be used as the parameter to this command.
%%\acmSubmissionID{123-A56-BU3}

%%
%% The majority of ACM publications use numbered citations and
%% references. The command \citestyle{authoryear} switches to the
%% If you are preparing content for an event
%% sponsored by ACM SIGGRAPH, you must use the "author year" style of
%% the next command will enable that style.

%%
%% end of the preamble, start of the body of the document source.
\sloppy

\title[Convolutional Transformer based Dual Discriminator GAN]{Convolutional Transformer based Dual Discriminator Generative Adversarial Networks for Video Anomaly Detection}

\author{Xinyang Feng}
\affiliation{
\institution{Columbia University}
\city{New York}
\state{New York}
\country{USA}
}
\email{xf2143@columbia.edu}

\author{Dongjin Song}\authornote{Corresponding author}
\affiliation{
\institution{University of Connecticut}
\city{Storrs}
\state{Connecticut}
\country{USA}
}
\email{dongjin.song@uconn.edu}

\author{Yuncong Chen}
\affiliation{
\institution{NEC Laboratories America, Inc.}
\city{Princeton}
\state{New Jersey}
\country{USA}
}
\email{yuncong@nec-labs.com}

\author{Zhengzhang Chen}
\affiliation{
\institution{NEC Laboratories America, Inc.}
\city{Princeton}
\state{New Jersey}
\country{USA}
}
\email{zchen@nec-labs.com}

\author{Jingchao Ni}
\affiliation{
\institution{NEC Laboratories America, Inc.}
\city{Princeton}
\state{New Jersey}
\country{USA}
}
\email{jni@nec-labs.com}

\author{Haifeng Chen}
\affiliation{
\institution{NEC Laboratories America, Inc.}
\city{Princeton}
\state{New Jersey}
\country{USA}
}
\email{haifeng@nec-labs.com}

\renewcommand{\shortauthors}{Xinyang Feng, Dongjin Song, Yuncong Chen, Zhengzhang Chen, Jingchao Ni, Haifeng Chen}

\begin{abstract}
% % \vspace{4mm}
 Detecting abnormal activities in real-world surveillance videos is an important yet challenging task as the prior knowledge about video anomalies is usually limited or unavailable. Despite that many approaches have been developed to resolve this problem, few of them can capture the normal spatio-temporal patterns effectively and efficiently. Moreover, existing works seldom explicitly consider the local consistency at frame level and global coherence of temporal dynamics in video sequences. To this end, we propose Convolutional Transformer based Dual Discriminator Generative Adversarial Networks (CT-D2GAN) to perform unsupervised video anomaly detection. Specifically, we first present a convolutional transformer to perform future frame prediction. It contains three key components, \textit{i.e.}, a convolutional encoder to capture the spatial information of the input video clips, a temporal self-attention module to encode the temporal dynamics, and a convolutional decoder to integrate spatio-temporal features and predict the future frame. Next, a dual discriminator based adversarial training procedure, which jointly considers an image discriminator that can maintain the local consistency at frame-level and a video discriminator that can enforce the global coherence of temporal dynamics, is employed to enhance the future frame prediction. Finally, the prediction error is used to identify abnormal video frames. Thoroughly empirical studies on three public video anomaly detection datasets, \textit{i.e.}, UCSD Ped2, CUHK Avenue, and Shanghai Tech Campus, demonstrate the effectiveness of the proposed adversarial spatio-temporal modeling framework.
\end{abstract}

\begin{CCSXML}
<ccs2012>
 <concept>
 <concept_id>10010147.10010178.10010224.10010225.10011295</concept_id>
 <concept_desc>Computing methodologies~Scene anomaly detection</concept_desc>
 <concept_significance>500</concept_significance>
 </concept>
 <concept>
 <concept_id>10010147.10010257.10010258.10010261.10010276</concept_id>
 <concept_desc>Computing methodologies~Adversarial learning</concept_desc>
 <concept_significance>500</concept_significance>
 </concept>
 <concept>
 <concept_id>10010147.10010257.10010258.10010260.10010229</concept_id>
 <concept_desc>Computing methodologies~Anomaly detection</concept_desc>
 <concept_significance>500</concept_significance>
 </concept>
 <concept>
 <concept_id>10010147.10010257.10010293.10010294</concept_id>
 <concept_desc>Computing methodologies~Neural networks</concept_desc>
 <concept_significance>500</concept_significance>
 </concept>
 </ccs2012>
\end{CCSXML}

\ccsdesc[500]{Computing methodologies~Scene anomaly detection}
\ccsdesc[500]{Computing methodologies~Adversarial learning}
\ccsdesc[500]{Computing methodologies~Anomaly detection}
\ccsdesc[500]{Computing methodologies~Neural networks}

\keywords{Video anomaly detection; Generative adversarial networks; Transformer model; Convolutional neural network; Spatio-temporal modeling}

\maketitle

\section{Introduction}
% \vspace{4mm}
With the rapid growth of video surveillance data, there is an increasing demand to automatically detect abnormal video sequences in the context of large-scale normal (regular) video data. Despite a substantial amount of research effort has been devoted to this problem~\cite{ucsdped_cvpr,ucsdped,cuhk,temporalregularity,xu2017detecting,shtech_cvpr,pred_recon,park2020cvpr,chang2020eccv}, video anomaly detection, which aims to identify the activities that do not conform to regular patterns in a video sequence, is still a challenging task. This is because real-world abnormal video activities can be extremely diverse while the prior knowledge about these anomalies is usually limited or even unavailable.
% In addition, the definitions of normal and abnormal activities are ambiguous in many cases. 

With the assumption that a model can only generalize to data from the same distribution as the training set, abnormal activities in the test set will manifest as deviance from regular patterns. A common approach to resolve this problem is to learn a model that can capture regular patterns in the normal video clips during the training stage, and check whether there exists any irregular pattern that diverges from regular patterns in the test video clips. Within this framework, it is crucial to not only represent the regular appearances but also capture the normal spatio-temporal dynamics to differentiate abnormal activities from normal activities in a video sequence. This serves as an important motivation for our proposed methods.

Early studies have used handcrafted features to represent video patterns~\cite{ucsdped_cvpr,ucsdped,cuhk,TIP2010}. For instance, Li et al.~\shortcite{ucsdped} introduced mixtures of dynamic textures and defined outliers under this model as anomalies. These approaches, however, are usually not optimal for video anomaly detection since the features are extracted based upon a different objective.

Recently, deep neural networks are becoming prevalent in video anomaly detection, showing superior performance over handcrafted feature based methods. For instance, Hasan et al.~\shortcite{temporalregularity} developed a convolutional autoencoder (Conv-AE) to model the spatio-temporal patterns in a video sequence simultaneously with a 2D CNN. The temporal dynamics, however, are not explicitly considered. To better cope with the spatio-temporal information in a video sequence, convolutional long short-term memory (LSTM) autoencoder (ConvLSTM-AE)~\cite{convlstm,convlstm_anomaly} was proposed to model the spatial patterns with fully convolutional networks and encode the temporal dynamics using convolutional LSTM~(ConvLSTM). ConvLSTM, however, suffers from computational and interpretation issues. A powerful alternative for sequence modeling is the self-attention mechanism~\cite{attention}. It has demonstrated superior performance and efficiency in many different tasks, \textit{e.g.}, sequence-to-sequence machine translation~\cite{attention}, time series prediction~\cite{IJCAI2017}, autoregressive model based image generation~\cite{image_xfmer}, and GAN-based image synthesis~\cite{sagan}. However, it has seldom been employed to capture regular spatio-temporal patterns in the surveillance videos.

More recently, adversarial learning has shown impressive progress on video anomaly detection. For instance, Ravanbakhsh et al.~\shortcite{gan_anomaly} developed a GAN based anomaly detection approach following conditional GAN framework~\cite{i2i}. Liu et al.~\shortcite{shtech_cvpr} proposed an anomaly detection approach based on future frame prediction. Tang et al.~\shortcite{pred_recon} extended this framework by adding a reconstruction task. The generative models in these two works were based on U-Net~\cite{unet}. Similar to Conv-AE, the temporal dynamics in the video clip were not explicitly encoded and the temporal coherence was enforced by a loss term on the optical flow.
Moreover, the potential discriminative information in the form of consistency at frame-level and global coherence of temporal dynamics in video sequences were not fully considered in previous works. 

In this paper, to better capture the regular spatio-temporal patterns and cope with the potential discriminative information at frame-level and in video sequences, we propose Convolutional Transformer based Dual Discriminator Generative Adversarial Networks (CT-D2GAN) to perform unsupervised video anomaly detection. We first present a convolutional transformer to perform future frame prediction. The convolutional transformer is essentially a 
encoder-decoder framework consisting of three key components, \textit{i.e.}, a \textit{convolutional encoder} to capture the spatial patterns of the input video clip, a novel \textit{temporal self-attention module} adapted for video temporal modeling that can explicitly encode the temporal dynamics, and a \textit{convolutional decoder} to integrate spatio-temporal features and predict the future frame. Because of the temporal self-attention module, convolutional transformer can capture the underlying temporal dynamics efficiently and effectively.
Next, in order to maintain the local consistency of the predicted frame and the global coherence conditioned on the previous frames, we adapt dual discriminator GAN to deal with video frames and employ an adversarial training procedure to further enhance the prediction performance. Finally, the prediction error is adopted to identify abnormal video frames. Thoroughly empirical studies on three public video anomaly detection datasets, \textit{i.e.}, UCSD Ped2, CUHK Avenue, and Shanghai Tech Campus, demonstrate the effectiveness of the proposed framework and techniques.

% \vspace{6mm}
\section{Related Work}
% \vspace{6mm}
The proposed Convolutional Transformer based Dual Discriminator Generative Adversarial Networks (CT-D2GAN) is closely related to deep learning based video anomaly detection and self-attention mechanism \cite{attention}.

% \subsection{Video Anomaly Detection based on Deep Neural Networks}
Note that we focus our discussions on methods based on unsupervised settings, which are efficient in generalization without the time-consuming and error-prone process of manual labeling. We are aware that there are numerous works on weakly supervised or supervised video anomaly detection, \textit{e.g.}, Sultani et al. (\citeyear{ucf}) proposed a deep multiple instance ranking framework using video-level labels and achieves better performance than convolutional auto-encoder (Conv-AE) based method~\cite{temporalregularity}, but it employs both normal and \emph{abnormal} video clips for training which is different from our setting.

Deep neural networks based video anomaly detection methods demonstrate superior performance over traditional methods based on handcrafted features. Hasan et al. (\citeyear{temporalregularity}) developed Conv-AE method to simultaneously learn the spatio-temporal patterns in a video with 2D convolutional neural networks by concatenating the video frames in the channel dimension. The temporal information is mixed with the spatial information in the first convolutional layer, thus not explicitly encoded. Xu et al. (\citeyear{xu2017detecting}) proposed appearance and motion DeepNet (AMDN) to learn video feature representations, which however still requires a decoupled one-class SVM classifier applied on learned representation to generate anomaly score. Dong et al. (\citeyear{memAE}) proposed a memory-augmented autoencoder (MemAE) that uses a memory module to constrain the reconstruction.

More recently, adversarial learning has demonstrated flexibility and impressive performance in multiple video anomaly detection studies. A generative adversarial networks (GANs) based anomaly detection approach~\cite{gan_anomaly} was developed following cGAN framework of image-to-image translation ~\cite{i2i}. Specifically, it employs image and optical flow as source domain and target domain, and vice versa, and trains cross-channel generation through adversarial learning. The reconstruction error is used to compute anomaly score. The only temporal constraint is imposed by the optical flow calculation.
Liu et al. (\citeyear{shtech_cvpr}) proposed an anomaly detection approach based on future frame prediction in GAN framework and U-Net~\cite{unet}. Similar to Conv-AE, the temporal information is not explicitly encoded and the temporal coherence between neighboring frames is enforced by a loss term on the optical flow. Tang et al. (\citeyear{pred_recon}) extended the future frame prediction framework by adding a reconstruction task.
One way to alleviate the temporal encoding issue in video spatio-temporal modeling is to use convolutional LSTM autoencoder (ConvLSTM-AE) based methods \cite{convlstm,chong2017abnormal,convlstm_anomaly, AAAI2019}, where the spatial and temporal patterns are encoded with fully convolutional networks and convolutional LSTM, respectively. Despite its popularity, ConvLSTM suffers from issues such as large memory consumption. The complex gating operations add to the computational cost and complicate the information flow, making interpretation difficult.

A more effective and efficient alternative for sequence modeling is the self-attention mechanism ~\cite{attention},
which is essentially an attention mechanism relating different positions of a single sequence to compute a representation of the sequence, in which the keys, values, and queries are from the same set of features.
Some related applications include autoregressive model based image generation~\cite{image_xfmer}, GAN-based image synthesis~\cite{sagan}.

In this work, based on related works, we introduce the convolutional transformer by extending the self-attention mechanism to video sequence modeling and develop a novel self-attention module specialized for spatio-temporal modeling in video sequences. Compared to existing approaches for video anomaly detection, the proposed convolutional transformer model has the advantage of being able to explicitly and efficiently encode the temporal information in a sequence of feature maps, where the computation of attentions can be fully parallelized via matrix multiplications. Based on the convolutional transformer, a dual discriminator generative adversarial networks (D2GAN) approach is developed to further enhance the future frame prediction through enforcing local consistency of the predicted frame and the global coherence conditioned on the previous frames. Note that the proposed D2GAN differs from existing works on dual discriminator based GAN which have been applied to different scenarios~\cite{d2,ddcgan,image_inpainting,access}.

\begin{figure*}[!t]
\centering
\includegraphics[width=18cm]{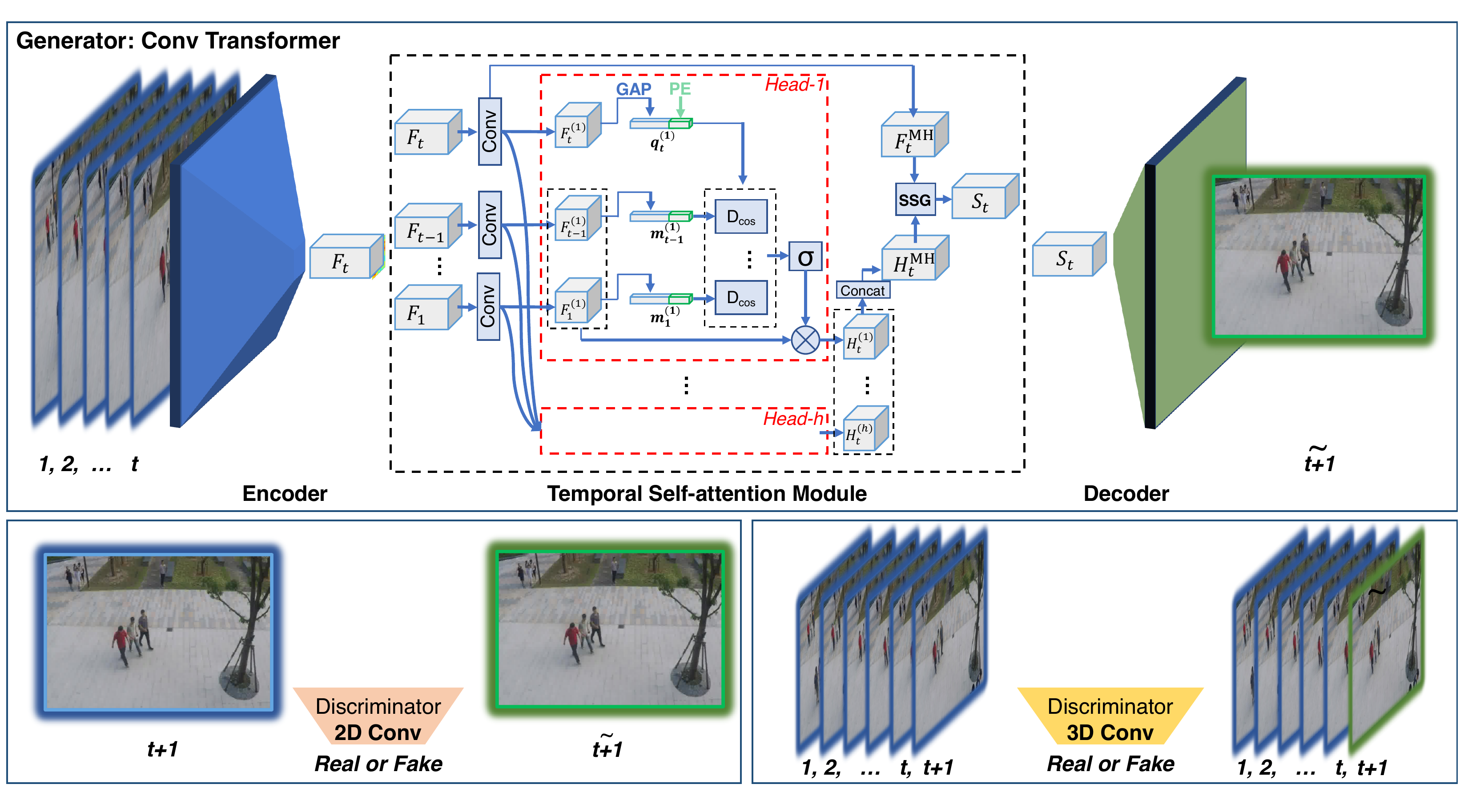}
\caption{The architecture of the proposed CT-D2GAN framework.
\textbf{(Upper panel)} The convolutional transformer generator is consisted of a convolutional encoder, a temporal self-attention module, and a convolutional decoder.
Multi-head self-attention is applied on the feature maps $F_t$ extracted from convolutional encoder: $F_t$ is transformed to multi-head feature maps $F_t^{\mathrm{(k)}}$ via a convolutional operation; within each head, we apply a global average pooling (GAP) operation on $F_{t}^{(k)}$ to generate a spatial feature vector by aggregating over spatial dimension, and concatenate the positional encoding (PE) vector; we then compare the similarity $D_{\mathrm{cos}}$ between query $\mathbf{q}_{t}^{(k)}$ and memory $\mathbf{m}_{t}^{(k)}$ feature vectors and generate the attention weights by normalizing across time steps using softmax $\sigma$; the attended feature map $H_t^{(h)}$ is a weighted average of the feature maps at different time steps; the final attended map $H_t^{\mathrm{MH}}$ is the concatenation over all the heads; the final integrated map $S_t$ is a weighted average of the query $F_t^{\mathrm{MH}}$ and the attended feature maps according to a spatial selective gate (SSG). $S_t$ is decoded to the predicted future frame with the convolutional decoder.
 \textbf{(Lower panels)} The image discriminator (left) and video discriminator (right) used in our dual discriminator GAN framework.
 }
\label{label:framework}% \vspace{-2mm}
\end{figure*}

\section{CT-D2GAN}

In this section, we first introduce the problem formulation and input to our framework. Then, we present the motivation and technical details of the proposed CT-D2GAN framework including convolutional transformer, dual discriminator GAN, the overall loss function, and lastly the regularity score calculation. An overview of the framework is illustrated in Figure \ref{label:framework}.

In CT-D2GAN, a convolutional transformer is employed to generate future frame prediction based on past frames, an image discriminator and a video discriminator are used to maintain the local consistency and global coherence.

\subsection{Problem Statement}
Given an input representation of video clip of length $T$, \textit{i.e.}, $I = (I_{t-T+1},...,I_{t}) \in \mathbb{R}^{h\times w\times c\times T}$, where $h$, $w$, $c$ are the height, width and number of channels, we aim to predict the $(t+1)$-th frame as $\hat{I}_{t+1} \in \mathbb{R}^{h\times w\times c}$ and identify abnormal activities based upon the prediction error, \textit{i.e.}, $e_{\text{MSE},t} = \frac{1}{h\cdot w\cdot c}\sum_{i=1}^{c}\|\hat{I}_{:,:,i,t+1} - I_{:,:,i,t+1}\|_F^{2}$, where $I_{:,:,i,t+1}\in \mathbb{R}^{h\times w}$.

\subsection{Input}
As appearance and motion are two characteristics of video data, it is common to explicitly incorporate optical flow together with the still images to describe a video sequence~\cite{two_stream}, \textit{e.g.} optical flow has been employed to represent video sequences in the cGAN framework~\cite{gan_anomaly} and used as a motion constraint~\cite{shtech_cvpr}.

In this work, we stack image with pre-computed optical flow maps~\cite{opticalflow,flownet2} in channel dimension as inputs similar to Simonyan et al.~\shortcite{two_stream} for video action recognition and Ravanbakhsh et al.~\shortcite{gan_anomaly} for video anomaly detection.
The optical flow maps consist of a horizontal component, a vertical component and a magnitude component. 
To be noted that, the optical flow map is computed from the previous image and current image, thus does not contain future frame information.
Therefore, the input can be given as $I\in \mathbb{R}^{h\times w \times 4 \times T}$, and we used 5 consecutive frames as inputs, \textit{i.e.}, $T=5$, similar to Liu et al.~\shortcite{shtech_cvpr}.

\subsection{Convolutional Transformer}
Convolutional transformer is developed to obtain a future frame prediction based on past frames. It consists of three key components: a convolutional encoder to encode spatial information, a temporal self-attention module to capture the temporal dynamics, and a convolutional decoder to integrate spatio-temporal features and predict future frame.

\subsubsection{Convolutional Encoder}
The convolutional encoder~\cite{fcn} is employed to extract spatial features from each frame of the video. Each frame of the video is first resized to $256\times256$ and then fed into the convolutional encoder. The convolutional encoder consists of 5 convolutional blocks. And the convolutional block follows common structure in CNN.
All the convolutional kernels are set as $3\times3$ pixels. For brevity, we denote a convolutional layer with stride $s$ and number of filters $n$ as Conv$_{s,n}$, a batch normalization layer as BN, a scaled exponential linear unit~\cite{selu} as SELU, and a dropout operation with dropout ratio $r$ as dropout$_r$. The structure of the convolutional encoder is: [Conv$_{1,64}$-SELU-BN]-[Conv$_{2,64}$-SELU-BN-Conv$_{1,64}$-SELU]-[Conv$_{2,128}$-SELU-BN-Conv$_{1,128}$-SELU]-[Conv$_{2,256}$-SELU-BN-dropout$_{0.25}$-Conv$_{1,256}$-SELU]-[Conv$_{2,256}$-SELU-BN-dropout$_{0.25}$-Conv$_{1,256}$-SELU] , where each [$\cdot$] represents a convolutional block. 

At the $l$-th convolutional block conv$^l$, $F_{t-i}^l\in\mathbb{R}^{h_l \times w_l \times c_l}, i \in [0,...,T-1]$ denotes the input feature maps to the self-attention module with $h_l$, $w_l$, $c_l$ as the height, width, and number of channels, respectively.
The temporal dynamics among the spatial feature maps of different time steps
will be encoded with temporal self-attention module.

\subsubsection{Temporal Self-attention Module}
\label{sec:sa}
To explicitly encode the temporal information in the video sequence, we extend self-attention mechanism in the transformer model~\cite{attention} and develop a novel temporal self-attention module to capture the temporal dynamics of the multi-scale spatial feature maps generated from the convolutional encoder. This section applies to all layers, thus we omit the layer for clarity.
An illustration of the multi-head temporal self-attention module is shown in the upper panel of Figure \ref{label:framework}. % 
% \vspace{1.5mm}
\noindent
\textbf{Spatial Feature Vector.}
We first use global average pooling (GAP) to extract a feature vector $\mathbf{f}_{t}$ from the feature map $F_{t}$ extracted in the convolutional encoder.
The feature vector in current time step $\mathbf{f}_{t}$ will be used as part of the query and each historical feature vector $\mathbf{f}_{t-i}$, $i\in[1,T-1]$ will be used as part of the key to index spatial feature maps. 

% \vspace{1.5mm}
\noindent
\textbf{Positional Encoding.}
Different from sequence models such as LSTM, self-attention does not model sequential information inherently, therefore it is necessary to incorporate temporal positional information into the model. We generate a positional encoding vector $\mathbf{PE} \in \mathbb{R}^{d_p}$ following~\cite{attention}:
\begin{equation}\label{eq:1}
 \begin{aligned}
\mathbf{PE}_{p,2i} = \mathrm{sin}(p/10000^{2i/d_p})
\\
\mathbf{PE}_{p,2i+1} = \mathrm{cos}(p/10000^{2i/d_p})
\end{aligned}
\end{equation}
\noindent where $d_p$ denotes the dimension of $\mathbf{PE}$, $p$ denotes the temporal position 
and $i\in [0,...\ ,(d_p/2-1)]$ denotes the index of the dimension. Empirically, we fix $d_p=8$ in our study.

% \vspace{1.5mm}
\noindent
\textbf{Temporal Self-Attention.}
\label{self-atten}
We concatenate the positional encoding vector with the spatial feature vector for each time step and use the concatenated vectors as the queries and keys, and the feature maps as the values in the setting of self-attention mechanism. For each query frame at time $t$, the current concatenated feature vector $\mathbf{q}_t =[\mathbf{f}_{t}; \mathbf{PE}]\in \mathbb{R}^{c_l+d_p}$ is used as query, and compared to the feature vector of each frame from the input video clip \textit{i.e.} memory $\mathbf{m}_{t-i}=[\mathbf{f}_{t-i}; \mathbf{PE}] \in \mathbb{R}^{c_l+d_p}, i \in [1,...,T-1]$ using cosine similarity:
\begin{equation}\label{eq:3}
D(\mathbf{q}_t,\mathbf{m}_{t-i}) = \frac{\mathbf{q}_t\cdot \mathbf{m}_{t-i}}{\|\mathbf{q}_t\| \|\mathbf{m}_{t-i}\|}.
\end{equation}
Based on the similarity between $\mathbf{q}_t$ and $\mathbf{m}_{t-i}$, we can generate the normalized attention weights $a_{t,i} \in \mathbb{R}$ across the temporal dimension using a softmax function:
\begin{equation}\label{eq:2}
a_{t,t-i} = \frac{\mathrm{exp}(\beta D(\mathbf{q}_t, \mathbf{m}_{t-i}))}{\sum_{j\in [1...T-1]} \mathrm{exp}(\beta D(\mathbf{q}_t, \mathbf{m}_{t-j}))},
\end{equation}
where a positive temperature variable $\beta$ is introduced to sharpen the level of focus in the softmax function and is automatically learned in the model through a single hidden densely-connected layer with the query as the input.

The final attended feature maps $H_t$ are a weighted sum of all feature maps $F$ 
using the attention weights in Eq. (\ref{eq:2}):
\begin{equation}\label{eq:HA}
\begin{split}
H_t &= \sum_{i\in [1,...,T-1]} a_{t,t-i} \cdot F_{t-i}.
\end{split}
\end{equation}

% \vspace{1.5mm}
\noindent
\textbf{Multi-head Temporal Self-Attention.}
Multi-head self-attention~\cite{attention} enables the model to jointly attend to information from different representation subspaces at different positions.
We adapt it to spatio-temporal modeling by first mapping the spatial feature maps to $n_h=8$ groups, each using 32 $1\times1$ convolutional kernels. For each group of feature maps with dimension $c_h=32$, we then perform the single head self-attention as described in the previous subsection and generate attended feature maps for head $k$ as $H_{t}^{(k)}$:
\begin{equation}\label{eq:ha_mh}
H_{t}^{(k)} = \sum_{i\in [1,...,T-1]} a_{t,t-i}^{(k)} \cdot F_{t-i}^{(k)},
\end{equation}
where $F_{t-i}^{(k)} \in \mathbb{R}^{ h_l \times w_l \times c_h}$ is the transformed feature map at frame $t-i$ for head $k$, $a_{t,t-i}^{(k)}$ is the corresponding attention weight.
The final multi-head attended feature map $H_t^{\mathrm{MH}}\in \mathbb{R}^{h_l \times w_l \times (c_h\cdot n_h)}$ is the concatenation of the attended feature maps from all the heads along the channel dimension:
\begin{equation}\label{eq:concat}
H_t^{\mathrm{MH}} = \mathrm{Concat}(H_{t}^{(1)},\ ...\ ,H_{t}^{(n_h)}).
\end{equation}
%\noindent
In this way, the final attended feature maps not only integrate spatial information from the convolutional encoder, but also capture temporal information from multi-head temporal self-attention mechanism.

% \vspace{1.5mm}
\noindent
\textbf{Spatial Selective Gate.}
The aforementioned module extends the self-attention mechanism to the temporal modeling of 2D image feature maps,
however, it comes with the loss of fine-grained spatial resolution due to the GAP operation. To compensate this, we introduce spatial selective gate (SSG), which is a spatial attention mechanism to integrate the current and historical information. The attended feature maps from the temporal self-attention module and the feature maps of the current query are concatenated, on which we learn a spatial selective gate using a sub-network $\mathcal{N}_{\mathrm{SSG}}$ with structure: Conv$_{1,256}$-BN-SELU-Conv$_{1,256}$-BN-SELU-Conv$_{1,256}$-BN-SELU-Conv$_{1,256}$-Conv$_{1,256}$-Sigmoid.
The final output is a pixel-wise weighted average of the attended maps $H_t^{\mathrm{MH}}$ and the current query's multi-head transformed feature maps $F_{t}^{\mathrm{MH}}\in \mathbb{R}^{h_l \times w_l \times (c_h\cdot n_h)}$, according to $SSG$:
\begin{equation}\label{eq:ssg}
S_t = SSG \circ F_{t}^{\mathrm{MH}} + (1-SSG) \circ H_t^{\mathrm{MH}}
\end{equation}
where $\circ$ denotes element-wise multiplication.

We add SSG at each level of temporal self-attention module. As the spatial dimensions are larger at shallow layers and we want to include contextual information while preserving the spatial resolution, we use dilated convolution~\cite{dilatedconvolution} with different dilatation factors at the 4 convolutional blocks in the sub-network $\mathcal{N}_{\mathrm{SSG}}$, specifically from conv$^2$ to conv$^5$, the dilation factors are (1,2,4,1), (1,2,2,1), (1,1,2,1), (1,1,1,1).
Note that SSG is computationally more efficient than directly forwarding the concatenated feature maps to the convolutional decoder.

\subsubsection{Convolutional Decoder}
The outputs of the temporal self-attention module $S_t$
are fed into the convolutional decoder. The convolutional decoder predicts the video frame using 4 transposed convolutional layers with stride 2 on the feature maps in a reverse order of the convolutional encoder. The fully-scaled feature maps then go through one convolutional layer with 32 filters and one convolutional layer with $c$ filters of size $1\times1$ that maps to the same size of channels $c$ in the input.
In order to predict finer details, we utilize the skip connection~\cite{unet} to connect the spatio-temporally integrated maps at each level of the convolutional encoder to the corresponding level of the convolutional decoder, which allows the model to further fine-tune the predicted frames.

\subsection{Dual Discriminator GAN}
We propose a dual discriminator GAN using both an image discriminator and a video discriminator to further enhance the future frame prediction of convolutional transformer via adversarial training. The image discriminator $D_I$ critiques on whether the current frame is generated or real just on the basis of one single frame to assess the local consistency. The video discriminator $D_V$ performs critique on the prediction conditioned on the past frames to assess the global coherence. Specifically, we stack the past frames with current generated or real frame in the temporal dimension and the video discriminator is essentially a video classifier.
This idea of combining local and global (contextual) discriminator is similar to adversarial image inpainting~\cite{image_inpainting} but is used in a totally different context.

The network structures of the two discriminators are kept the same except that we use 2D operations in image discriminator and the corresponding 3D operations in the video discriminator.
We use PatchGAN architecture as described in~\cite{i2i} and use spectral normalization~\cite{spectralnorm} in each convolutional layer. In the 3D version, the stride and kernel size in the temporal dimension were set at 1 and 2 respectively.

The method in Liu et al.~\shortcite{shtech_cvpr} is similar to using the image discriminator only.
Different from the video discriminator in Tulyakov et al.~\shortcite{mocogan}, which applies on the whole synthetic video clip, our proposed video discriminator conditions on the real frames.

\begin{table*}[!htbp]
\centering
\caption{Video anomaly detection datasets details}
\label{dataset}
\def\arraystretch{1.3}
\rotatebox{0}{
\begin{tabular}{c|c|c|c|c}
\hline
Dataset & Total \# frames/clips & Training \# frames/clips & Testing \# frames/clips & Anomaly Types \\
\hline
UCSD Ped2 & 4,560/28 & 2,550/16 & 2,010/12 & biker, skater, vehicle\\
\hline
CUHK Avenue & 30,652/37 & 15,328/16 & 15,324/21 & running, loitering, object throwing\\
\hline
{ShanghaiTech} & {315,307/437} & {274,516/330} & {40,791/107} & biker, skater, vehicle, sudden motion \\
\hline
\end{tabular}
}
\end{table*}

\subsection{Loss}
For the adversarial training, we use the Wasserstein GAN with gradient penalty (WGAN-GP) setting~\cite{wgan,wgangp}.
The generator $G$ is the mapping
: $I \rightarrow \widetilde{I}_{t+1}$. For discriminators, $D_V: (I, \hat{I}_{t+1}) \rightarrow p[(I,\hat{I}_{t+1})\ \text{is real}] $ and $D_I: \hat{I}_{t+1} \rightarrow p[\hat{I}_{t+1}\ \text{is real}]$ are video and image discriminators respectively.
The GAN loss is:
\begin{equation}\label{eq:gan}
\begin{split}
L_{adv}(G, D_I, D_V) & = \mathbb{E}_{I,\widetilde{I}_{t+1}}[D_V(I,\widetilde{I}_{t+1})] - \mathbb{E}_{I,I_{t+1}}[D_V(I,I_{t+1})] \\
 & + \lambda \mathbb{E}_{I,\hat{I}_{t+1}}[(\| \nabla D_V(I,\hat{I}_{t+1}) \|_2 - 1)^2] \\
 & + \mathbb{E}_{\widetilde{I}_{t+1}}[D_I(\widetilde{I}_{t+1})] - \mathbb{E}_{I_{t+1}}[D_I(I_{t+1})] \\
 & + \lambda \mathbb{E}_{\hat{I}_{t+1}}[(\| \nabla D_I(\hat{I}_{t+1}) \|_2 - 1)^2] \\
\end{split}
\end{equation}
where $\hat{I}_{t+1} = \epsilon {I}_{t+1} + (1-\epsilon)\widetilde{I}_{t+1}$, $\epsilon \sim U[0,1]$.
The penalty coefficient $\lambda$ is fixed as 10 in all our experiments. 

In addition, we consider the pixel-wise $L_1$ loss of the prediction. Therefore the total loss $L$ is:
\begin{equation}\label{eq:totalloss}
\begin{split}
L = L_{adv} + \|{I}_{t+1} - \widetilde{I}_{t+1}\|_1
\end{split}
\end{equation}
We trained our models on each dataset separately by minimizing the loss above using ADAM~\cite{adam} algorithm with learning rate 0.0002 and a batch size of 5.

\subsection{Regularity Score}
A regularity score based on the prediction error $e_t$ is calculated for each video frame:
\begin{equation}\label{eq:error}
r_{e_t} = 1 - \frac{e_t-\mathrm{min}_{\tau}e_{\tau}}{\mathrm{max}_{\tau} e_{\tau} -\mathrm{min}_{\tau}e_{\tau}}
\end{equation}
In Hasan et al.~\shortcite{temporalregularity},
$e_t$ is the frame-wise reconstruction $e_{\text{MSE},t}$.
In Liu et al.~\shortcite{shtech_cvpr},
$e_t$ is equivalently negative frame-wise prediction PSNR (Peak Signal to Noise Ratio): PSNR $=10\mathrm{log}_{10}\frac{\mathrm{max}(\widetilde{I}_t)}{e_{\text{MSE},t}}$.
In this study, we use similar setting to the two methods above with: $e_t=\mathrm{log}_{10}e_{\text{MSE},t}$.

\section{Experiments}
In this section, we first introduce the three public datasets used in our experiments, which follow the same setup as other similar unsupervised video anomaly detection studies. Then, we report the video anomaly detection performance and comparison with other methods. Finally, we perform ablation studies to demonstrate the contribution of each component and interpret the results based on the proposed CT-D2GAN.

\subsection{Datasets}
We evaluate our framework on three widely used public video anomaly detection datasets, 
\textit{i.e.}, UCSD Ped2 dataset~\cite{ucsdped} 
\footnote{\small{\url{http://www.svcl.ucsd.edu/projects/anomaly/dataset.html}}},
CUHK Avenue dataset~\cite{cuhk}
\footnote{\small{\url{http://www.cse.cuhk.edu.hk/leojia/projects/detectabnormal/dataset.html}}},
and ShanghaiTech Campus (SH-Tech) dataset~\cite{shtech}
\footnote{\small{\url{https://github.com/StevenLiuWen/sRNN_TSC_Anomaly_Detection##shanghaitechcampus-anomaly-detection-dataset}}}.
We describe the dataset-specific characteristics and the effects on video anomaly detection performance, some details can be found in Table \ref{dataset}:
\subsubsection{UCSD Ped2.} UCSD Ped2 includes pedestrians, vehicles largely moving in parallel to the camera plane.
\subsubsection{CUHK Avenue.} CUHK Avenue includes pedestrians and objects both moving parallel to or toward/away from the camera. Slight camera motion is present in the dataset. Some of the anomalies are staged actions.
\subsubsection{ShanghaiTech.} Different from the other datasets, the ShanghaiTech dataset is a multi-scene dataset (13 scenes), and includes pedestrians, vehicles, and sudden motions, and the ratios of each scene in the training set and test set can be different.

\begin{figure*}[!hbtp]
\centering
\includegraphics[width=17.8cm]{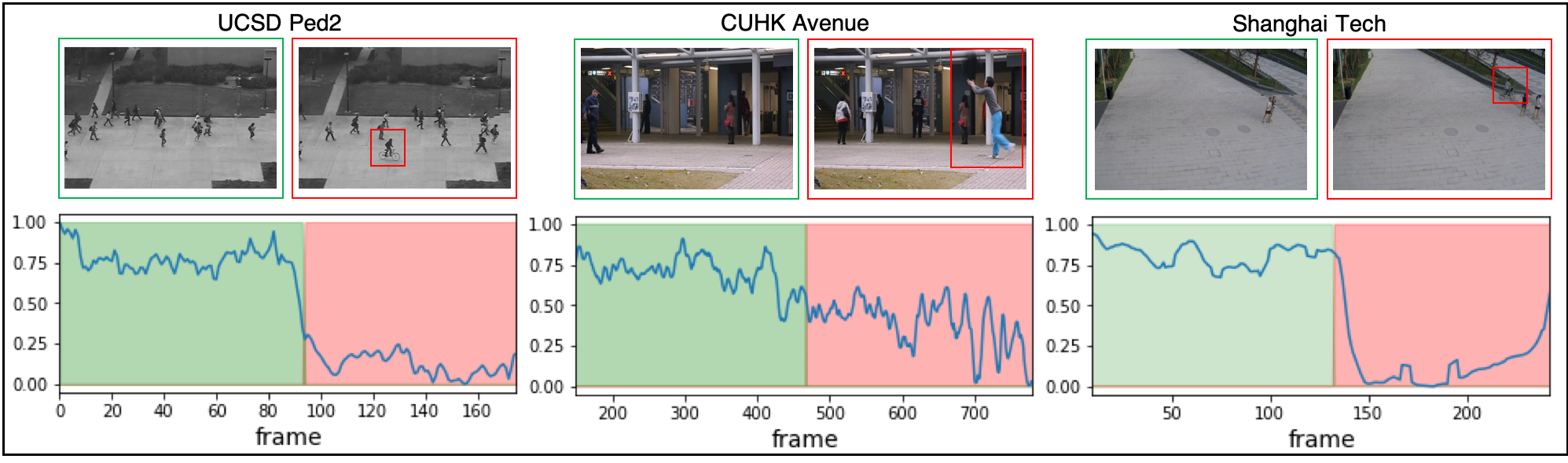}
\caption{Examples of video anomaly detection. The blue lines in the line graphs delineate frame-level regularity scores. The green and red shaded segments in the line graphs indicate the ground truth normal and abnormal video segments respectively. The frames in the green boxes are regular frames from the regular video segments, the frames in the red boxes are abnormal frames from abnormal video segments. The abnormal objects are annotated.}
\label{score}
\end{figure*}

% \subsection{Evaluations}
% \subsubsection{Anomaly Detection}
\subsection{Evaluation}
The model was trained and evaluated on a system with an NVIDIA GeForce 1080 Ti GPU and implemented with PyTorch. To measure the effectiveness of our proposed CT-D2GAN framework for video anomaly detection, 
we report the area under the receiver operating characteristics (ROC) curve \textit{i.e.}, AUC. Specifically, AUC is calculated by comparing the frame-level regularity scores with frame-level ground truth labels.

\newcounter{reftbl}
\begin{table}[!hbtp]
\centering
\caption{Frame-level video anomaly detection performance (AUC).}
\label{performance}
\def\arraystretch{1.2}
\begin{tabular}{c|c|c|c}
\hline
Method & UCSD Ped2 & CUHK & SH-Tech\\
\hline
\hline
MPPCA+SF~\shortcite{ucsdped_cvpr} & 61.3 & - & - \\
\hline
MDT~\shortcite{ucsdped_cvpr,ucsdped} & 82.9 & - & - \\
\hline
\hline
Conv-AE~\shortcite{temporalregularity} &\ 85.0 $^{\dagger}$& \ 80.0 $^{\dagger}$ &\ 60.9 $^{\dagger}$ \\
\hline
3D Conv~\shortcite{spatio-temporal-ae} & 91.2& 80.9 & - \\
\hline
Stacked RNN~\shortcite{shtech} & 92.2 & 81.7 & 68.0 \\
\hline
ConvLSTM-AE~\shortcite{convlstm_anomaly} & 88.1 & 77.0 & - \\
\hline
memAE~\shortcite{memAE} & 94.1 & 83.3 & 71.2 \\
\hline
memNormality~\shortcite{park2020cvpr} & 97.0 & \textbf{88.5} & 70.5 \\
\hline
ClusterAE~\shortcite{chang2020eccv} & 96.5 & 86.0 & 73.3 \\
\hline
\hline
AbnormalGAN~\shortcite{gan_anomaly} & 93.5 & - & -\\
\hline
Frame prediction~\shortcite{shtech_cvpr} & 95.4 & 85.1 & 72.8 \\
\hline
Pred+Recon~\shortcite{pred_recon} & 96.3 & 85.1 & 73.0 \\
\hline
% & & & & & &\\
\hline
CT-D2GAN & \textbf{97.2} & 85.9 & \textbf{77.7} \\
\hline
\end{tabular}

% \end{tabular*}

\begin{quote}

$^{\dagger}$ Evaluated in~\cite{shtech_cvpr};

-: Not evaluated in the study.

Ordered in publication year.
The best performance in each dataset is highlighted in \textbf{boldface}.

\end{quote}
\end{table}

\subsection{Video Anomaly Detection}
To demonstrate the effectiveness of our proposed CT-D2GAN framework for video anomaly detection, we compare it against 12 different baseline methods. Among those, MPPCA (mixture of probabilistic principal component analyzers) + SF (social force)~\cite{ucsdped_cvpr}, MDT (mixture of dynamic textures)~\cite{ucsdped_cvpr,ucsdped}
are handcrafted feature based methods;
Conv-AE~\cite{temporalregularity}, 3D Conv~\cite{spatio-temporal-ae}, Stacked RNN~\cite{shtech}, and ConvLSTM-AE~\cite{convlstm_anomaly} are encoder-decoder based approaches; MemAE~\cite{memAE}, MemNormality~\cite{park2020cvpr} and ClusterAE~\cite{chang2020eccv} are recent encoder-decoder based methods enhanced with memory module or clustering;
AbnormalGAN~\cite{gan_anomaly}, Frame prediction~\cite{shtech_cvpr}, and Pred+Recon~\cite{pred_recon} are methods based on adversarial training. 

Table \ref{performance} shows the frame-level video anomaly detection performance (AUC) of various approaches.
We observed that encoder-decoder based approaches in general outperform handcrafted feature based methods. This is because the handcrafted features are usually extracted based upon a different objective and thus can be sub-optimal. Within encoder-decoder based approaches, ConvLSTM-AE outperforms Conv-AE since it can better capture temporal information.
We also notice that adversarial training based methods perform better than most baseline methods.
Finally, our proposed CT-D2GAN framework achieves the best performance on UCSD Ped2 and SH-Tech, and close to the best performance in CUHK~\cite{park2020cvpr}. This is because our proposed model can not only capture the spatio-temporal patterns explicitly and effectively through convolutional transformer but also leverage the dual discriminator GAN based adversarial training to maintain local consistency at frame-level and global coherence in video sequences.
Recent memory or clustering enhanced methods ~\cite{park2020cvpr,chang2020eccv,memAE} show good performance and is orthogonal to our proposed framework and can integrate with our proposed framework in future work to further improve performance.
Examples of video anomaly detection results overlaid on the abnormal activity ground truth of all three datasets are shown in Figure \ref{score}, along with example video frames from the regular and abnormal video segments. 

Due to the multi-scene nature of SH-Tech dataset, we also analyzed the most ample single scene that constitutes 25\% (83/330 clips) of training set and 32\% (34/107 clips) of test set, the AUC is 87.5 which is much better than the overall dataset and reach similar level with other single-scene datasets. This could imply that generalizing to less ample scenes is still a challenging task given unbalanced training set.

Thanks to the convolutional transformer architecture and optimizations including spatial selective gate, our model is computationally efficient. At inference time, our model runs at 45 FPS on one NVIDIA GeForce 1080 Ti GPU.

\begin{table}[hbtp]
\centering
 
\caption{Video anomaly detection performance under different ablation settings on UCSD Ped2 dataset.}
\label{table:ablation}
\def\arraystretch{1.1}
\begin{tabular}{c|c}
\hline
Ablation setting & AUC \\
\hline
\hline
Conv Transformer & 94.2 \\
\hline
Conv Transformer + image discriminator & 95.7 \\
\hline
Conv Transformer + video discriminator & 96.9 \\
\hline
U-Net + dual discriminator & 95.5 \\
\hline
CT-D2GAN & \textbf{97.2} \\
\hline
\end{tabular}
\end{table}

\subsection{Ablation Studies}
To understand how each component contributes to the anomaly detection task, we conducted ablation studies with different settings: (1) convolutional transformer only without the adversarial training (Conv Transformer), (2) Conv Transformer with image discriminator only, (3) Conv Transformer with video discriminator only, (4) U-Net based generator (as has been utilized in image-to-image translation~\cite{i2i} and video anomaly detection~\cite{shtech_cvpr}) with dual discriminator, and compare with our full CT-D2GAN model. 
The performance comparison can be found in Table \ref{table:ablation}. We observed that adversarial training can enhance the performance for anomaly detection, either with the image discriminator or the video discriminator. Video discriminator alone achieves nearly similar performance as using dual discriminator, but we observed the loss decreased faster when combined with image discriminator. Using image discriminator alone was not as effective, and the loss was less stable. Finally, we observed that CT-D2GAN achieved superior performance than U-Net with dual discriminator, suggesting that 
convolutional transformer can better capture the spatio-temporal dynamics and thus can make a more accurate detection.

\subsection{Interpretation}
\begin{figure}[t]
\centering
\includegraphics[width=8.5cm]{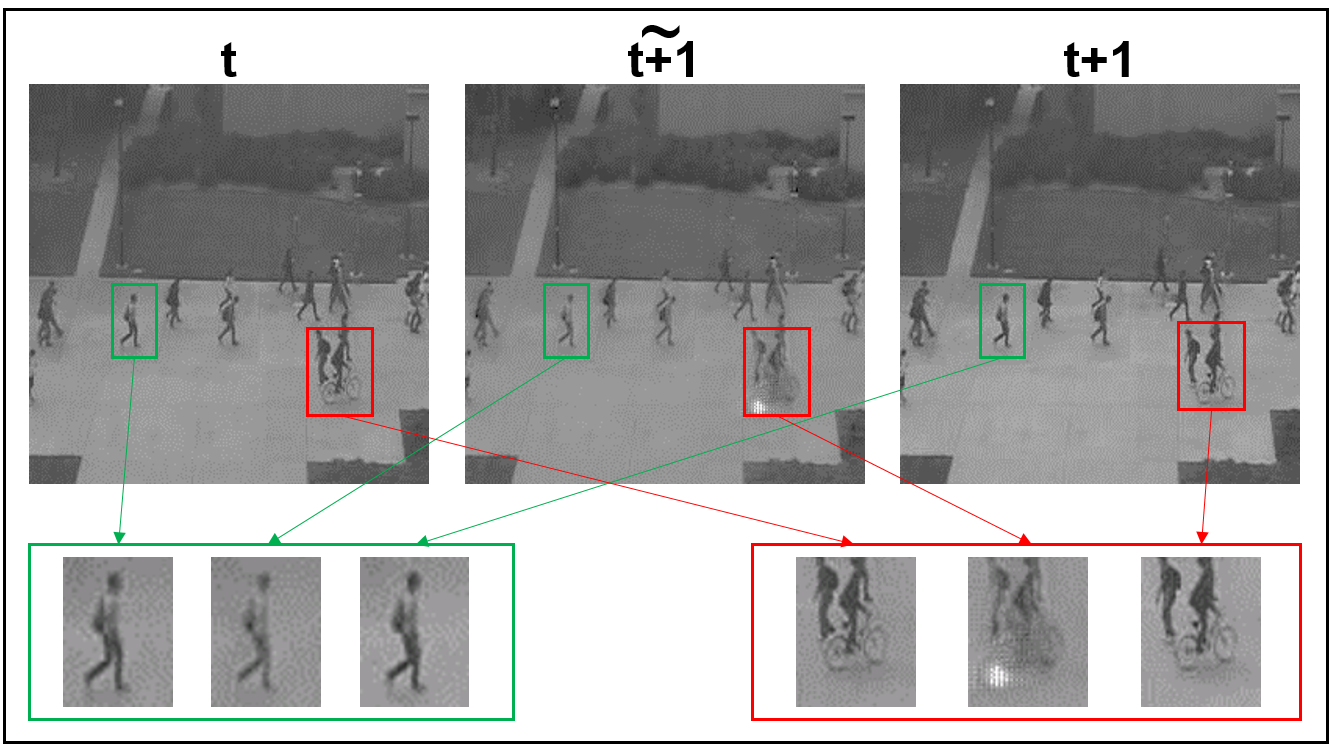}
\caption{ An example showing the future frame prediction in the normal part of the image (green box, pedestrian in this case) where we observe the model capturing the dynamics of the behavior, and abnormal part of the image (red box, bicycle in this case) where there is large prediction error. From left to right, we show the last frame in the input video clip ($t$), the predicted future frame $\widetilde{t+1}$, and the ground truth future frame $t+1$.}
\label{pred}% \vspace{-4mm}
\end{figure}

We illustrate an example of predicted future frame $\widetilde{t+1}$ and compare it with the previous frame $t$ and the ground truth future frame $t+1$ in Figure \ref{pred}. The prediction performance is poor for the anomaly (red box). And also we noted that the model is able to capture the temporal dynamics by predicting the future behavior in normal part of the image (green box).

\begin{figure}[t]
\centering
\includegraphics[width=8.5cm]{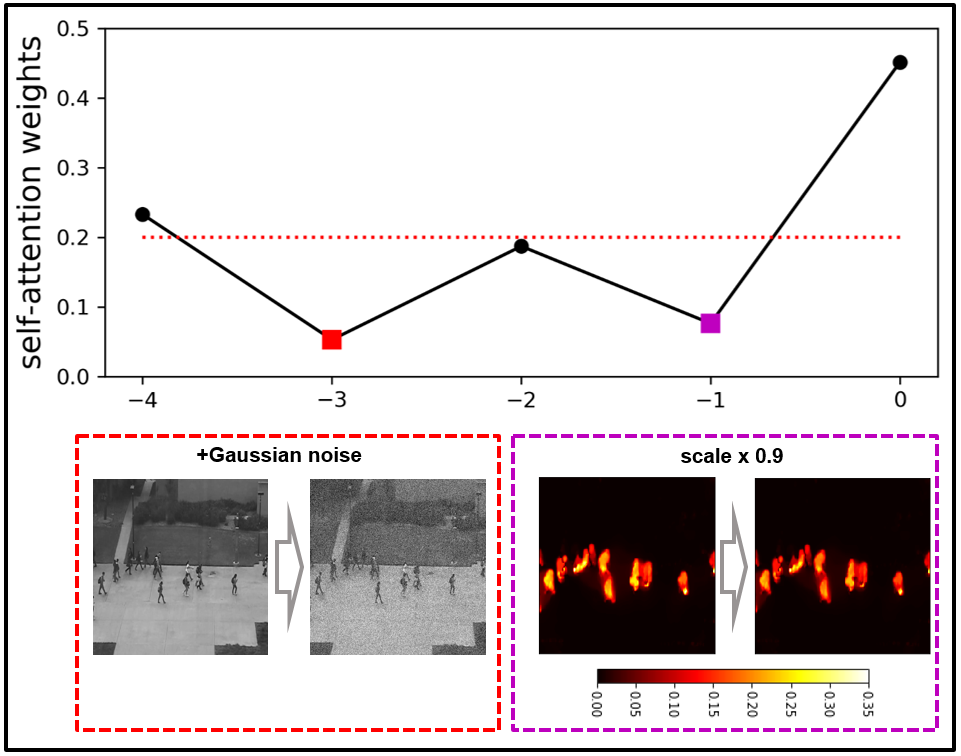}
\caption{ Temporal self-attention weights in perturbed video clip.}
\label{atten}% \vspace{-4mm}
\end{figure}

\noindent
\textbf{Self-attention weights under perturbation.}
It is not straightforward to directly interpret the temporal self-attention weight vector, as temporal self-attention is applied to an abstract representation of the video. Therefore, to further investigate the effectiveness of temporal self-attention, we perturb two frames of the video and run the inference on this perturbed video segment. For one frame (Figure \ref{atten}, red), we added a random Gaussian noise with zero mean and 0.1 standard deviation to the image to simulate the deterioration in video quality; for another frame (Figure \ref{atten}, purple), we scaled the optical flow maps by 0.9 to simulate the frame rate distortion. We plot the temporal attention weights for the frame right after the two perturbed frames in Figure ~\ref{atten}. The weights assigned to the perturbed frames are clearly lower than the others, implying less contribution to the attended map. This suggests that self-attention module can adaptively select relevant feature maps and is robust to input noise. 

\section{Conclusions}

In this paper, we developed Convolutional Transformer based Dual Discriminator Generative Adversarial Networks (CT-D2GAN) to perform unsupervised video anomaly detection.
The convolutional transformer which consists of three components, \textit{i.e.}, a convolutional encoder to capture the spatial patterns of the input video clip, a temporal self-attention module to encode the temporal dynamics, and a convolutional decoder to integrate spatio-temporal features, was employed to perform future frame prediction. A dual discriminator based adversarial training approach was used to maintain the local consistency of the predicted frame and the global coherence conditioned on the previous frames. Thorough experiments on three widely used video anomaly detection datasets demonstrate that our proposed CT-D2GAN is able to detect anomaly frames with superior performance.

% \vspace{3mm}

% \begin{acks}
% \end{acks}

\bibliographystyle{ACM-Reference-Format}
\balance
\bibliography{safcn_gan_anomaly_detection_draft}
\end{document}